\documentclass[runningheads]{llncs}

\usepackage{accv}

\usepackage{accvabbrv}

\usepackage{graphicx}
\usepackage{booktabs}
\usepackage{lipsum}
\usepackage{multirow}

\usepackage[accsupp]{axessibility}  

\usepackage{hyperref}

\usepackage{orcidlink}

\newif\ifshowrevisions
\showrevisionsfalse
\DeclareRobustCommand{\revised}[1]{\ifshowrevisions\textcolor{blue}{#1}\else#1\fi}
\newcommand{\revisioncolor}{\ifshowrevisions\color{blue}\fi}
\begin{document}

\title{CHASE-VLA: Post-Training Quantization Framework for Vision-Language-Action Models with Chunk-Aware Scale Estimation} 

\titlerunning{CHASE-VLA: PTQ Framework for VLA Models}

\author{Jin Hyun\inst{1,2}\orcidlink{0000-0002-4038-0170} \and
Jung Gyu Min\inst{1,2}\orcidlink{0009-0002-2041-617X} \and
Gyuhyun Jung\inst{1,2}\orcidlink{0009-0001-4826-537X} \and
Youngjoo Lee\inst{1}\orcidlink{0000-0002-2467-8276}}

\authorrunning{J.~Hyun et al.}

\institute{KAIST, Daejeon, South Korea\\
\email{youngjoo@kaist.ac.kr}
\and
POSTECH, Pohang, South Korea\\
\email{\{hyunj, mjg1104, jgh1020\}@postech.ac.kr}}
\maketitle

\begin{abstract}
Vision-Language-Action (VLA) models map visual observations and language instructions to continuous robot actions, but a diffusion-based action expert (AE) poses a key challenge for low-bit post-training quantization (PTQ).
The AE is repeatedly invoked across denoising steps and policy queries, where fixed calibration scales can be mismatched with activation ranges that vary with denoising progress and intended motion.
We propose CHASE-VLA, a chunk-aware PTQ method that exploits a VLA-specific signal readily available from the policy: the generated action chunk, including its unexecuted future suffix.
Rather than relying only on static scale matching for AE layers, CHASE-VLA combines the previously generated chunk as causal action context with denoising step group information to adapt AE activation scales.
This enables W4A4 quantization of both MLP and attention projections in the repeated AE without modifying the pretrained policy.
On LIBERO, CHASE-VLA achieves 97.3\% average success rate on $\pi_{0.5}$ when both MLP and attention projections in the AE are quantized to W4A4, restoring FP16-level performance.
CHASE-VLA also reduces the weight storage of the quantized AE linear layers by 73.4\% and their single-chunk memory traffic by 70.9\% and 71.2\% on $\pi_{0.5}$ and GR00T N1.6, respectively, with a predictor overhead of at most 1.26\% of the saved storage.
\end{abstract}

\section{Introduction}
\label{sec:intro}

Vision-Language-Action (VLA) models have emerged as a foundation for embodied intelligence by connecting vision-language understanding with executable robot control \cite{brohan2023rt1,brohan2023rt2,kim2024openvla,li2024cogact,black2024pi0,bjorck2025gr00t}.
By integrating visual perception, language understanding, high-level reasoning, and action generation within a single policy, these models allow robots to interpret natural language instructions, ground them in visual observations, and produce control commands for manipulation tasks \cite{brohan2023rt2,kim2024openvla,team2024octo,li2024cogact,black2024pi0,kim2025openvlaoft}.
Previous VLA systems typically treat action generation as a lightweight output interface, where robot actions are discretized into language-compatible tokens and predicted through the Vision-Language Model (VLM) output head rather than a dedicated action-generation module \cite{brohan2023rt1,brohan2023rt2,kim2024openvla,bu2025univla,wang2025vqvla}.

\begin{figure}[tb]
  \centering
  \includegraphics[height=6cm]{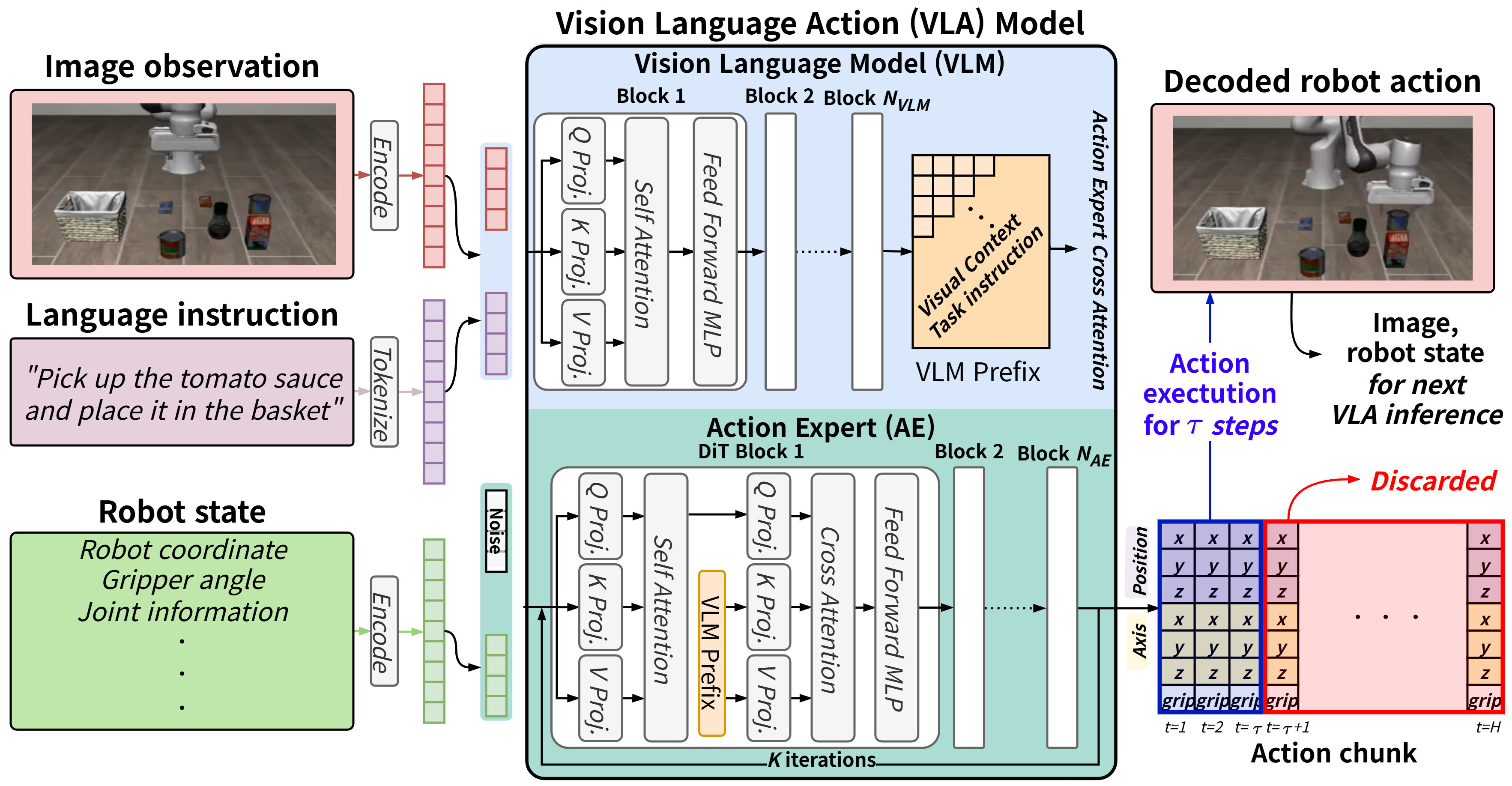}
  \caption{VLA architecture with a diffusion-based AE for action generation.}
  \label{fig:example1}
\end{figure}

In contrast, recent VLA architectures increasingly preserve the perception and reasoning capability of the VLM branch while introducing a specialized diffusion-based action expert (AE) for continuous trajectory generation \cite{chi2023diffusionpolicy,team2024octo,li2024cogact,liu2024rdt1b,black2024pi0,wen2025dexvla,wen2025diffusionvla}.
This transition makes the AE a central component of modern VLA policies, because it directly determines trajectory smoothness, temporal consistency, and control robustness during robot execution \cite{li2024cogact,black2024pi0,wen2025dexvla,zhang2026quantvla,xu2026qvla,xu2026daptq}.
Fig.~\ref{fig:example1} summarizes this architectural trend: the VLM branch provides vision-language grounding, while a separate diffusion-based AE repeatedly refines continuous action chunks for robot control.

Unlike action-token decoders, a diffusion-based AE generates action chunks through iterative refinement, repeatedly processing transformer blocks across denoising steps \cite{chi2023diffusionpolicy,peebles2023scalable,li2024cogact}.
This computation recurs at every replanning step throughout a rollout, so weight reads and activation movement accumulate over the entire trajectory \cite{zhang2026quantvla}.
The AE is therefore a primary target for deployment-oriented quantization because it directly affects both memory traffic and control fidelity in repeated action generation \cite{xu2026qvla}.
Among compression options, post-training quantization (PTQ) is a practical approach for pretrained policies because it avoids the robot data collection, rollout validation, and iterative action generation required by quantization-aware training \cite{nagel2020adaround,jacob2018quantization,wang2025bitvla}.
This makes low-bit PTQ attractive for deployment, but the quantized AE must still preserve continuous-control precision across repeated denoising and replanning throughout the executed trajectory \cite{xu2026qvla,zhang2026quantvla}.

\begin{figure}[tb]
  \centering
  \includegraphics[width=\linewidth]{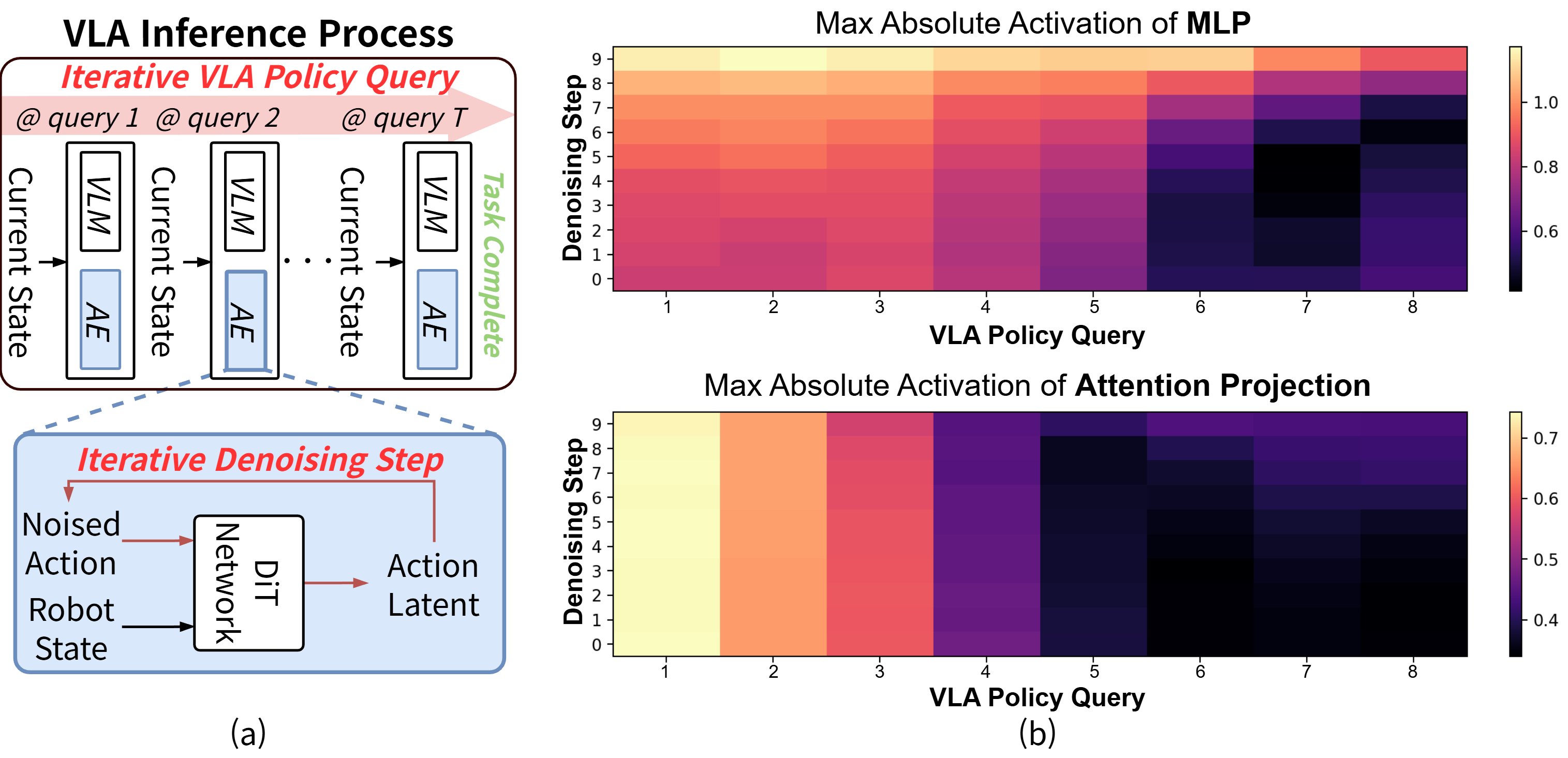}
  \caption{Activation-scale variation in a diffusion-based VLA AE. (a) Repeated AE calls across denoising steps and policy queries. (b) Maximum absolute activation values across denoising steps and policy queries, which make fixed calibration scales prone to clipping or loss of quantization resolution.}
  \label{fig:action_expert_distribution}
\end{figure}

Low-bit PTQ for the AE therefore differs from language-model quantization in both sensitivity and execution pattern \cite{xiao2023smoothquant,frantar2023gptq,lin2024awq,ashkboos2024quarot,zhang2026quantvla}.
Previous VLA PTQ methods keep sensitive attention projections in floating point and quantize only the MLP projections \cite{zhang2026quantvla,xu2026daptq}, leaving a large fraction of the repeated AE cost unquantized.
This limitation stems from a mismatch between static calibration scales and the dynamic activation behavior of the repeatedly executed AE \cite{ptq4dit,qdit,viditq,zhang2026quantvla}.
As illustrated in Fig.~\ref{fig:action_expert_distribution}(a), the AE is invoked in two nested loops: it refines each action chunk over multiple denoising steps and regenerates a new chunk at every replanning step throughout a rollout \cite{chi2023diffusionpolicy,black2024pi0,li2024cogact}.
Across denoising steps, the AE produces physical control values through iterative refinement, so its activations change with the denoising step rather than following a fixed forward-pass distribution \cite{zhang2026quantvla}.
The activation range also changes with the action context.
Each policy query is conditioned on a different observation, instruction, and robot state, so the generated motion, and therefore the required activation scale, can vary substantially from one chunk to the next \cite{black2024pi0,li2024cogact,xu2026daptq}.
Fig.~\ref{fig:action_expert_distribution}(b) shows this behavior within individual AE linear layers, where the maximum absolute activation varies across both denoising steps and action contexts.
A single calibration-derived scale cannot accommodate these variations without either clipping broad activations or losing resolution for fine control \cite{zhang2026quantvla,zheng2026dyqvla}.
Across repeated denoising and replanning, the resulting local errors can accumulate into trajectory-level control errors \cite{xu2026daptq}.
Addressing this mismatch requires activation scales that depend on both the denoising step and the motion being generated \cite{ptq4dit,qdit,zheng2026dyqvla}.
However, the action context of the current chunk is not causally available before the AE produces that chunk, making direct current-chunk scale selection impractical.

Chunk-based VLA execution provides a causal source of action context for resolving this causality constraint.
Modern VLA policies predict short action chunks at each policy query instead of a single immediate action \cite{chi2023diffusionpolicy,li2024cogact,black2024pi0, bjorck2025gr00t}.
Each chunk constitutes a short-horizon plan conditioned on the current observation, language instruction, and robot state.
As illustrated in Fig.~\ref{fig:example1}, the robot executes only an initial prefix of the chunk and then replans after receiving a new observation \cite{chi2023diffusionpolicy,black2024pi0, bjorck2025gr00t}.
The remaining suffix is not executed because the next policy query generates a new chunk based on the updated observation.
However, this suffix is still the policy's predicted continuation of the same trajectory whose prefix is executed.
Thus, it provides an accessible signal of the intended near-future motion without additional rollouts or task-specific annotations.

We propose CHASE-VLA, a chunk-aware PTQ method for low-bit quantization of diffusion-based VLA AEs.
Instead of using fixed calibration scales, CHASE-VLA modulates calibrated activation scales with action-conditioned factors that are shared within groups of denoising steps.
The conditioning signal is the previously generated action chunk, which compactly summarizes the intended near-future motion and enables scale selection to reflect both action context and denoising progress without modifying the AE architecture.
Through adaptive scaling, CHASE-VLA supports 4-bit weight and activation quantization (W4A4) and mitigates the scale mismatch that typically arises from static PTQ.
Unlike previous VLA PTQ methods, which keep sensitive attention projections in floating point for stability, CHASE-VLA extends W4A4 coverage beyond MLP layers to the attention projection layers within the repeated AE.
By including these repeatedly invoked projections in the low-bit quantization target, CHASE-VLA achieves further reductions in memory traffic compared to MLP-only AE quantization.

Our contributions are summarized as follows.
\begin{itemize}
    \item[$\bullet$] We identify activation-scale mismatch across denoising steps and action contexts as a central barrier to low-bit PTQ for diffusion-based VLA AEs.
    \item[$\bullet$] \revised{CHASE-VLA estimates activation scales from the previous action chunk and applies them at the denoising step group level.}
    \item[$\bullet$] CHASE-VLA expands the W4A4 AE target from MLP projections to both MLP and attention projections while keeping the pretrained policy and the AE architecture unchanged.
    \item[$\bullet$] Experiments on LIBERO across $\pi_{0.5}$, GR00T N1.6, and CogACT show that this design largely preserves task success under the expanded W4A4 target, while an efficiency analysis on $\pi_{0.5}$ and GR00T N1.6 quantifies reductions in quantized AE linear weights and single-chunk memory traffic with modest predictor overhead.
\end{itemize}

\section{Related Work}

\subsection{Vision-Language-Action Models}

\revised{Recent VLA research has shifted from discrete action prediction to continuous trajectories generated by diffusion-based AEs.}
Diffusion Policy demonstrated that modeling robot actions as a conditional denoising process is effective for generating smooth and multimodal action trajectories \cite{chi2023diffusionpolicy}.
Following this direction, recent systems such as CogACT, $\pi_{0.5}$, GR00T, and RDT employ diffusion-based action generation modules to couple high-level vision-language reasoning with low-level continuous control \cite{li2024cogact,black2024pi0,black2025pi05,bjorck2025gr00t,liu2024rdt1b}.
These AE architectures are attractive because they predict action chunks rather than isolated actions, allowing temporally correlated future controls to be generated as a coherent trajectory \cite{chi2023diffusionpolicy,black2025pi05, bjorck2025gr00t}.

The chunk-based formulation enhances trajectory smoothness and execution reliability, which are critical for contact-rich manipulation and long-horizon robotic tasks \cite{chi2023diffusionpolicy,li2024cogact,liu2024rdt1b,bjorck2025gr00t}.
However, this property introduces a substantial deployment burden as the AE repeatedly evaluates action chunks during multiple denoising steps.
The AE is therefore a major determinant of continuous-control quality and a key target for low-bit deployment \cite{zhang2026quantvla, xu2026daptq}.

\subsection{Efficient VLA Inference and Compression}
Recent studies have explored several approaches to reducing the inference cost of VLA models without sacrificing their embodied control capabilities.
\revised{TinyVLA combines compact multimodal transformers with lightweight diffusion-policy heads, while EfficientVLA accelerates inference by pruning redundant language layers and reusing intermediate representations \cite{wen2025tinyvla,efficientvla}.}
VLA-Cache further improves runtime efficiency by reusing cached key-value features for static visual tokens, and MoLe-VLA dynamically skips non-essential computation through mixture-of-layers routing in the language backbone \cite{xu2026vla,molevla}.
Most of these efforts primarily modify the model architecture or inference execution path, while the numerical precision of VLA policies remains less explored.
For lightweight deployment under tight memory and bandwidth budgets, this leaves quantization as an important requirement beyond architectural or execution-level optimization.
This requirement is particularly relevant to diffusion-based AEs, where low-bit inference must preserve action-chunk-dependent activation distributions and denoising step dynamics.


\subsection{Post-Training Quantization for LLMs, VLMs, DiTs, and VLAs}

For VLA AE quantization, a central distinction is not only quantization scope but also the information used for activation-scale selection during iterative action generation.
Table~\ref{tab:related_quantization} organizes this distinction across relevant PTQ categories and highlights the remaining gap addressed by CHASE-VLA.
\begin{table}[t]
\centering
\caption{Scale-selection signals and quantization scopes across PTQ categories.}
\label{tab:related_quantization}
\small
\renewcommand{\arraystretch}{1.18}
\setlength{\tabcolsep}{6pt}
\begin{tabular*}{\linewidth}{@{\extracolsep{\fill}}ccc@{}}
\toprule
PTQ Category & Scale-Selection Signal & Quantization Scope \\
\midrule
LLM/VLM PTQ & Weight/activation statistics & Transformer encoder/decoder \\
Diffusion PTQ & Denoising step signal & Diffusion denoising blocks \\
VLA PTQ & Static module calibration & AE MLP only \\
\textbf{CHASE-VLA} & Previous action chunk & AE MLP \\
\textbf{(Proposed)} & + denoising step group & + attention projections \\
\bottomrule
\end{tabular*}
\end{table}

\revised{PTQ reduces the memory footprint and inference cost of pretrained transformers without full retraining \cite{frantar2023gptq,xiao2023smoothquant}.}
For large language and vision-language models, representative methods such as SmoothQuant, GPTQ, and AWQ address low-bit deployment by reducing activation outliers, compensating weight quantization error, or redistributing activation energy through smoothing and scaling \cite{xiao2023smoothquant,frantar2023gptq,lin2024awq,ashkboos2024quarot}.
These methods provide useful building blocks for quantizing transformer backbones in VLA systems, where semantic grounding and multimodal reasoning are handled by the VLM branch.
However, their scale choices are mainly designed for token-generating transformer inference and do not explicitly model the iterative denoising dynamics or action-context variation of diffusion-based AEs.
Directly transferring these assumptions to VLA action generation can therefore be insufficient when quantization errors perturb continuous robot commands.

\revised{Quantization has also been studied for diffusion models and diffusion transformers (DiT), whose activation statistics vary across denoising steps \cite{ptq4dit,qdit}.}
Methods such as PTQ4DiT and Q-DiT improve diffusion quantization through denoising-step-aware calibration, fine-grained grouping, dynamic quantization, or low-rank residual compensation.
These methods are relevant to modern VLA policies because diffusion-based AEs also refine outputs over denoising steps.
Nevertheless, diffusion PTQ methods are primarily developed for image or video generation, where scale adaptation is usually tied to the denoising process itself.
They do not use the predicted action chunk as causal context for selecting activation scales in subsequent AE inference.
This leaves a gap between denoising-step-aware diffusion quantization and action-context-aware quantization for VLA AEs.

More recently, VLA-specific PTQ methods have begun to address the numerical sensitivity of VLA pipelines directly.
QuantVLA analyzes scale drift in DiT-based VLA AEs and introduces calibration-based stabilization for VLA PTQ \cite{zhang2026quantvla}.
Its AE quantization uses a selective target: attention projections are kept in floating point while the low-bit target focuses on MLP projections \cite{zhang2026quantvla}.
DA-PTQ addresses a complementary cross-space mismatch by aligning the conditioning representation passed from the vision-language backbone to the action decoder \cite{xu2026daptq}.
This conditioning-pathway alignment is distinct from adapting activation scales inside the repeatedly executed AE.

Together, these VLA-specific methods improve PTQ stability but still leave open the issue of W4A4 quantization across both MLP and attention projections within the AE.
Rather than relying only on static calibration statistics, denoising step cues, or cross-module representation alignment, CHASE-VLA uses the previously generated action chunk and denoising step groups as conditioning signals for activation-scale estimation.
This formulation directly targets the action-context- and denoising-step-dependent scale mismatch that arises during iterative action chunk generation.

\section{Method}

\begin{figure}[tb]
  \centering
  \includegraphics[width=\linewidth]{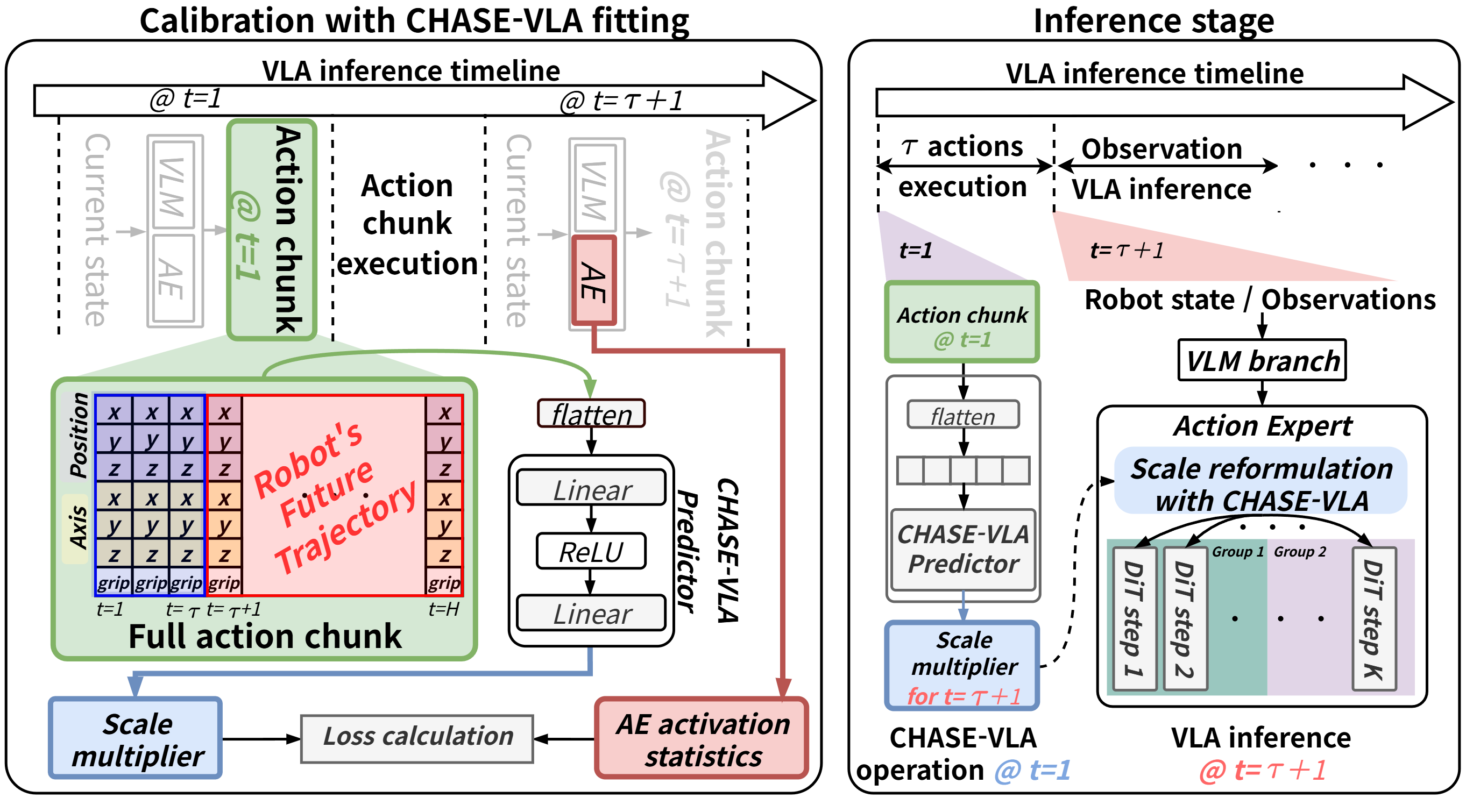}
  \caption{Overview of CHASE-VLA. After each policy query, the generated action chunk, including its unexecuted future suffix, provides causal action context for precomputing activation scales for denoising step groups in the next query.}
  \label{fig:example2}
\end{figure}

\subsection{Preliminaries and Static PTQ}

Modern VLA policies commonly pair a vision-language branch with a diffusion-based AE.
Let $t$ denote a control time at which the policy is queried, and let $\tau \le H$ be the number of actions executed before the next query.
Given an observation image $I_t$ and a language instruction $L$, the vision-language branch produces a conditioning representation $F_t=\mathrm{VLM}(I_t,L)$.
Together with the robot proprioceptive state $r_t$, this representation conditions an AE that iteratively refines an action latent and decodes it into an action chunk $A_t=[a_{t,0},\ldots,a_{t,H-1}]$, where $a_{t,h}\in\mathbb{R}^{d_a}$ denotes the $h$-th robot action in the chunk, $d_a$ is the action dimension, and $H$ is the action horizon.
At the AE denoising step $k\in\{0,\ldots,K-1\}$, where $K$ is the total number of denoising steps, let $x_k$ denote the action latent at step $k$.
The AE update is

\begin{equation}
x_{k}=f_\theta(x_{k+1},F_t,r_t,k),
\label{eq:action_expert_update}
\end{equation}
where $f_\theta$ is the denoising function.
The final denoised latent is mapped to the executable chunk by the action decoder,
\begin{equation}
A_t=\mathrm{Dec}(x_0).
\label{eq:action_chunk_decode}
\end{equation}
$\mathrm{Dec}$ denotes the final action decoder.

CHASE-VLA targets this repeated action-generation path.
Each target linear layer in $f_\theta$ is invoked across denoising steps and replanned chunks, so its memory traffic is repeatedly amplified during a rollout.
At the same time, numerical errors in this path directly perturb continuous robot actions rather than only semantic token probabilities.

For a target linear layer $l$, let $u_{t,l,k}$ denote its input activation tensor when the policy generates chunk $A_t$ at denoising step $k$.
Uniform PTQ quantizes this tensor as
\begin{equation}
\begin{aligned}
q(u;s) &= \operatorname{clip}(\operatorname{round}(u/s),q_{\min},q_{\max}),\\
\hat{u}(s) &= q(u;s)\cdot s.
\end{aligned}
\label{eq:static_quant}
\end{equation}
In this expression, $q(u;s)$ is the integer quantization of activation tensor $u$, $\hat{u}(s)$ is its dequantized value, $s$ denotes the activation scale, $q_{\max}=2^{a_{\mathrm{bits}}-1}-1$ and $q_{\min}=-q_{\max}$ for symmetric signed $a_{\mathrm{bits}}$-bit activations, and the operator is applied element-wise.
The scale $s$ determines the symmetric clipping range.

Static PTQ derives the activation scale of each target layer from a calibration set and keeps it fixed throughout inference.
In Eq.~\ref{eq:static_quant}, a fixed scale $s$ sets the clipping threshold to $q_{\max}\cdot s$.
If the actual activation range exceeds this threshold, activations are clipped; if it is much smaller, the quantization step becomes unnecessarily coarse, reducing resolution.
However, this range varies with the denoising step and the policy query, since each query generates a different action chunk.
A fixed scale, therefore, cannot track the per-query and per-step activation range, motivating an activation scale that depends on both the denoising step and the action context.

\subsection{Chunk-Aware Scale Estimation}
\label{sec:scale_adaptation}

To address this variation, activation-scale selection should consider both the denoising step and the specific action being generated.
A fixed calibration scale ignores both factors, while a denoising-step-only rule still assigns identical scales to different action chunks processed at the same denoising step.
As illustrated in Fig.~\ref{fig:example2}, CHASE-VLA overcomes these limitations by retaining the quantization operator in Eq.~\ref{eq:static_quant} while reusing the previously generated action chunk to precompute group-wise activation scales for generating the next action chunk.
Instead of predicting an absolute scale independently, CHASE-VLA predicts a multiplicative adjustment to a calibration-derived base scale, utilizing the generated action context and the denoising step group.


CHASE-VLA derives action context from action chunks previously predicted by the policy.
For a policy query at time $t$, the preceding query at $t-\tau$ generated a full action chunk $A_{t-\tau}$, and the robot executed only its initial $\tau$ actions before replanning.
Although the remaining suffix is unexecuted, it represents the policy's predicted continuation of the executed prefix and thus encodes the intended near-future motion.
Therefore, as shown in Fig.~\ref{fig:example2}, CHASE-VLA conditions the activation scale for the current chunk $A_t$ on the entire previous chunk, including both the prefix and the suffix.
This approach provides a signal that captures motion magnitude and temporal structure without requiring ground-truth future actions or rollout annotations.


The scale adjustment is also conditioned on the denoising step. 
Each adjustment is a multiplicative factor on the base scale.
Rather than assigning an independent factor to every step, CHASE-VLA maps the $K$ denoising steps to $M$ groups indexed by $m\in\{1,\ldots,M\}$ and predicts one factor per group.
This grouping keeps the predictor output compact while still distinguishing the differing activation ranges of coarse early and fine late refinements.

CHASE-VLA initially determines the base scale, which subsequently undergoes adjustment based on action-conditioned group parameters.
Here, $\tilde{u}_{l,m}$ represents the recorded calibration activations for layer $l$ corresponding to denoising step group $m$.
The group-specific initial scale is
\begin{equation}
  s^{\mathrm{init}}_{l,m}
  =
  \max\left(
  \epsilon_s,\,
  \frac{\operatorname{absmax}(\tilde{u}_{l,m})}{q_{\max}}
  \right),
\label{eq:initial_scale}
\end{equation}
where $\operatorname{absmax}(\cdot)$ takes the maximum absolute value over all tensors in the calibration activation set and $\epsilon_s$ is a small scale floor.
These initial scales are used when generating the first action chunk of an episode, where no previous action chunk is available, and they anchor the relative scale adjustments used for later queries.
For a policy query at time $t$, CHASE-VLA forms the action context by flattening the previously generated chunk,
\begin{equation}
z_t=\mathrm{vec}(A_{t-\tau})\in\mathbb{R}^{H d_a}.
\label{eq:chase_action_context}
\end{equation}
The conditioning process is causal since $z_t$ is computed prior to the generation of the current chunk $A_t$ by the AE.
As shown in Fig.~\ref{fig:example2}, this process establishes a recurrent scale update path.
The complete chunk generated at time $t-\tau$, including its unexecuted future suffix, is reused as context for selecting activation scales at time $t$.

For each target linear layer $l$, $f_{\phi_l}$ denotes the lightweight CHASE-VLA predictor shown in Fig.~\ref{fig:example2}.
This predictor maps the normalized action context to one log-scale adjustment and one uncertainty estimate per denoising step group:
\begin{align}
\left[\hat{y}_{t,l,1:M},\log\hat{\sigma}^{2}_{t,l,1:M}\right]
=
f_{\phi_l}(\operatorname{norm}(z_t)).
\label{eq:chase_predictor_all_groups}
\end{align}
The output $\hat{y}_{t,l,m}$ is the predicted log-scale adjustment for group $m$, $\log\hat{\sigma}^{2}_{t,l,m}$ is the predicted log-variance used for uncertainty estimation, and $\operatorname{norm}(\cdot)$ denotes feature standardization with calibration-set mean and standard deviation.
For denoising step group $m$, CHASE-VLA converts the predictor output into an uncertainty-aware scale multiplier,
\begin{align}
\gamma_{t,l,m}
=
10^{\hat{y}_{t,l,m}+\alpha\hat{\sigma}_{t,l,m}},
\label{eq:chase_gamma}
\end{align}
where $\hat{\sigma}_{t,l,m}=\exp(0.5\log\hat{\sigma}^{2}_{t,l,m})$ and $\alpha$ is a fixed margin.
The uncertainty term moderates the context-specific adjustment when the prediction is uncertain, reducing the risk of an overly aggressive scale reduction under action contexts that deviate from the calibration distribution.
The adaptive factor is shared within denoising step group $m$ and corresponds to the scale multiplier shown in Fig.~\ref{fig:example2}; it modulates the group-specific initial scale to produce the final group scale:
\begin{equation}
s^{\text{CHASE-VLA}}_{t,l,m}
=
s^{\mathrm{init}}_{l,m}\cdot\gamma_{t,l,m}.
\label{eq:chase_final_scale}
\end{equation}
During denoising, each denoising step in the AE utilizes the precomputed scale of its assigned group.
Eq.~\ref{eq:chase_final_scale} therefore defines the activation scale for the current query from two sources: the previous action chunk $A_{t-\tau}$ and the denoising step group of the current linear call.

To fit the predictor, each collected calibration query is treated as a supervised pair: the previous chunk provides the predictor input, and the recorded activation statistics from producing the corresponding current chunk provide the target.
For the $i$-th collected query at time $t_i$, the predictor input is the stored context $\mathrm{vec}(A_{t_i-\tau})$, while $\tilde{u}_{t_i,l,m}$ is the activation observed when the AE generated the current chunk $A_{t_i}$.
As shown in Fig.~\ref{fig:example2}, these observed activation statistics are used only to construct offline calibration targets, not during inference.
Unlike the query-independent initial scale in Eq.~\ref{eq:initial_scale}, the current-chunk scale is computed separately for each collected query and denoising step group:
\begin{equation}
s^{\mathrm{chunk}}_{t_i,l,m}=
\frac{\operatorname{absmax}(\tilde{u}_{t_i,l,m})}{q_{\max}}.
\label{eq:chase_chunk_scale}
\end{equation}
With a small numerical constant $\epsilon$, the group target is the base-10 log-ratio between the group-specific current-chunk scale and the group-specific initial scale
\begin{equation}
y^{*}_{t_i,l,m}
=
\log_{10}
\left(
\epsilon+
\frac{s^{\mathrm{chunk}}_{t_i,l,m}}{s^{\mathrm{init}}_{l,m}}
\right).
\label{eq:chase_training_target}
\end{equation}
This target trains CHASE-VLA to use the previous chunk of the $i$-th collected query to predict how much of the calibration-derived group-specific range is needed for the corresponding current chunk at $t_i$.
The learned adjustment therefore modulates the calibrated group-specific scale to recover quantization resolution when the base range is unnecessarily broad, rather than regressing an absolute activation scale from scratch.
The predictor is fitted on collected calibration queries with a lightweight objective combining a pinball loss on the log-scale target, a Gaussian negative log-likelihood term for the log-variance head, and small parameter regularization.
A detailed analysis of the loss function utilized for CHASE-VLA predictor is provided in the supplement.
Throughout this calibration, the pretrained VLA policy remains frozen, so CHASE-VLA performs post-training quantization rather than policy fine-tuning.

\subsection{Quantization Targets and Inference Path}

CHASE-VLA focuses W4A4 quantization on repeated linear projections inside the diffusion AE, where the same projections are invoked across denoising steps and policy queries.
The adaptive W4A4 target therefore includes feed forward MLP projections and linear projections in the attention modules of the AE.
These layers lie on the repeated denoising path, where activation-scale mismatch directly affects low-bit action generation.

This target coverage is more aggressive than that of previous VLA PTQ methods, which typically quantize only the MLP in the AE while leaving attention projections in floating point.
The difference is important because attention projections are also invoked at every denoising step; keeping them in floating point leaves a substantial repeated data-movement cost within the AE.
As described in Sec.~\ref{sec:scale_adaptation}, CHASE-VLA adapts the activation scale to the predicted action context and denoising step group, allowing attention projections to remain within the W4A4 AE target instead of being left in floating point.

As shown in Fig.~\ref{fig:example2}, CHASE-VLA acts as a scale update module for the AE during inference.
Since the previous chunk is not available for the first policy query, the target linear layers use the initial scales in Eq.~\ref{eq:initial_scale}.
Once the AE generates $A_t$, CHASE-VLA immediately maps $\mathrm{vec}(A_t)$ to group-wise scaling factors and updates the activation scales for all target layers and denoising step groups of the next policy query at time $t+\tau$.
Thus, before the AE starts generating $A_{t+\tau}$, the scales $s^{\mathrm{CHASE-VLA}}_{t+\tau,l,m}$ in Eq.~\ref{eq:chase_final_scale} have already been fixed.
During the next denoising process, each target linear layer only reads the precomputed scale for its denoising step group $m$.
After $A_{t+\tau}$ is produced, CHASE-VLA similarly updates the scales for the following query from $\mathrm{vec}(A_{t+\tau})$.
The pretrained policy weights and denoising schedule remain unchanged; only the activation scale used by each target linear layer is updated between policy queries.

\section{Experiments}
\label{sec:blind}

\subsection{Experimental Setup}
CHASE-VLA is evaluated on three VLA models, $\pi_{0.5}$, GR00T N1.6, and CogACT, which represent diverse AE architectures and model scales \cite{black2025pi05, bjorck2025gr00t,li2024cogact}.
All models are evaluated on the four LIBERO task suites: Spatial, Object, Goal, and Long \cite{liu2023libero}, and we report the task success rate averaged over 20 episodes for each task within a suite.

We compare CHASE-VLA with the FP16 reference, QuantVLA \cite{zhang2026quantvla}, and diffusion PTQ baselines \cite{qdit,ptq4dit} adapted to the AE.
All compared PTQ methods use the same calibration set and evaluation protocol within each model.
For the diffusion PTQ baselines, we apply channel-wise salience balancing and Spearman's $\rho$-guided salience calibration for PTQ4DiT, and automatic granularity allocation with sample-wise dynamic activation quantization for Q-DiT.
QuantVLA is implemented by following its calibration and static scale-matching procedure using attention temperature matching and output head balancing.
Each evaluation focuses on the AE, while the VLM linear layers are quantized using low-bit PTQ methods with the LLM-oriented PTQ scheme \cite{duquant}, matching the VLM treatment in QuantVLA \cite{zhang2026quantvla}, ensuring that the reported differences are due solely to AE quantization.
\revised{We also report AE latency, quantized linear layer storage and memory traffic, and predictor overhead.}

\subsection{Quantized VLA Performance}

\begin{table}[t]
\centering
\caption{LIBERO success rates on $\pi_{0.5}$.}
\label{tab:main_pi05}
\small
\setlength{\tabcolsep}{1.5pt}
\begin{tabular*}{\linewidth}{@{\extracolsep{\fill}}lcccccccc@{}}
\toprule
\multirow{2}{*}{Method} & \multirow{2}{*}{Precision} & \multicolumn{2}{c}{AE quant. target} & \multirow{2}{*}{Spa.} & \multirow{2}{*}{Obj.} & \multirow{2}{*}{Goal} & \multirow{2}{*}{Long} & \multirow{2}{*}{Avg.} \\
\cmidrule{3-4}
 & & MLP & Attn. proj. & & & & & \\
\midrule
FP16 & FP16 & -- & -- & 98.5 & 99.0 & 97.5 & 93.5 & 97.1 \\
\hline
QuantVLA & W4A8 & \checkmark & -- & 98.5 & 98.0 & 98.0 & 96.0 & 97.6 \\
QuantVLA (expanded) & W4A4 & \checkmark & \checkmark & 98.5 & 98.5 & 93.5 & 90.5 & 95.3 \\
 \hline
Q-DiT & W4A4 & \checkmark & \checkmark & 97.0 & 97.5 & 95.0 & 88.0 & 94.4 \\
PTQ4DiT & W4A4 & \checkmark & \checkmark & 97.5 & 94.5 & 94.0 & 88.0 & 93.5 \\
\hline
CHASE-VLA (w/o action) & W4A4 & \checkmark & \checkmark & 98.5 & 97.5 & 93.5 & 92.5 & 95.5 \\
CHASE-VLA & W4A4 & \checkmark & \checkmark & 97.0 & 99.5 & 98.0 & 94.5 & 97.3 \\
\bottomrule
\end{tabular*}
\end{table}

Table~\ref{tab:main_pi05} presents the LIBERO evaluation results on $\pi_{0.5}$.
\revised{QuantVLA is evaluated in its original W4A8 MLP-only configuration and on the same expanded W4A4 AE target as CHASE-VLA.}
\revised{CHASE-VLA (w/o action) removes previous-chunk conditioning while retaining denoising step groups.}

\revised{With the common W4A4 target, CHASE-VLA achieves 97.3\% average success, exceeding expanded QuantVLA, Q-DiT, and PTQ4DiT by 2.0, 2.9, and 3.8 percentage points.}
\revised{Its 95\% confidence interval (CI), [95.9, 98.2]\%, overlaps the FP16 interval, [95.7, 98.1]\%, with similar point estimates.}
\revised{Removing action context yields 95.5\%, a 1.8-point reduction, with lower Goal and Long success.}

\begin{table}[t]
\centering
\caption{LIBERO success rates on additional VLA models.}
\label{tab:main_crossmodel}
\small
\setlength{\tabcolsep}{4pt}
\begin{tabular}{@{}llcccccc@{}}
\toprule
Model & Method & Precision & Spa. & Obj. & Goal & Long & Avg. \\
\midrule
\multirow{5}{*}{CogACT} & FP16 & FP16 & 87.5 & 98.0 & 95.5 & 91.5 & 93.1 \\
 & QuantVLA (expanded) & W4A4 & 89.5 & 57.5 & 73.5 & 81.0 & 75.4 \\
 & Q-DiT & W4A4 & 81.0 & 73.5 & 83.5 & 73.5 & 77.9 \\
 & PTQ4DiT & W4A4 & 87.5 & 77.5 & 80.5 & 73.0 & 79.6 \\
 & CHASE-VLA & W4A4 & 86.5 & 97.0 & 88.0 & 90.5 & 90.5 \\
\midrule
\multirow{5}{*}{\shortstack{GR00T\\N1.6}} & FP16 & FP16 & 99.0 & 99.5 & 97.5 & 96.0 & 98.0 \\
 & QuantVLA (expanded) & W4A4 & 53.0 & 55.0 & 56.0 & 33.5 & 49.4 \\
 & Q-DiT & W4A4 & 6.5 & 69.5 & 12.0 & 3.5 & 22.9 \\
 & PTQ4DiT & W4A4 & 6.0 & 62.5 & 9.0 & 2.5 & 20.0 \\
 & CHASE-VLA & W4A4 & 90.5 & 99.5 & 92.0 & 84.0 & 91.5 \\
\bottomrule
\end{tabular}
\end{table}


\revised{Table~\ref{tab:main_crossmodel} extends the evaluation to CogACT and GR00T N1.6, where CHASE-VLA achieves 90.5\% and 91.5\% success, respectively.}
\revised{Their 95\% CIs, [88.3, 92.3]\% and [89.4, 93.2]\%, lie above those of all compared W4A4 PTQ baselines.}
\revised{The largest separation occurs on GR00T N1.6, where expanded QuantVLA, Q-DiT, and PTQ4DiT reach 49.4\%, 22.9\%, and 20.0\%.}
\revised{Detailed confidence interval results are reported in Supplementary.}

\subsection{\revised{Ablation Study}}
\revised{Table~\ref{tab:main_ablations} evaluates five design choices on $\pi_{0.5}$ across the four LIBERO suites, with the default CHASE-VLA configurations marked in bold.}

\begin{table}[t]
\revisioncolor
\centering
\caption{\revised{Ablation study of CHASE-VLA on $\pi_{0.5}$ across the four LIBERO suites. Success rates correspond to the settings in the same order; bold entries mark the default CHASE-VLA configuration. Prefix and suffix denote the executed and unexecuted portions of the previous action chunk.}}
\label{tab:main_ablations}
\small
\setlength{\tabcolsep}{2pt}
\begin{tabular*}{\linewidth}{@{\extracolsep{\fill}}lcc@{}}
\toprule
Factor & Tested settings & Mean success (\%) \\
\midrule
Denoising groups ($M$) & 1 / \textbf{3} / 5 / 10 & 95.5 / \textbf{97.3} / 97.1 / 97.4 \\
Uncertainty margin ($\alpha$) & 0 / 1 / \textbf{3} / 5 & 96.3 / 96.9 / \textbf{97.3} / 94.8 \\
Action context & Prefix / Suffix / \textbf{Full chunk} & 95.9 / 96.3 / \textbf{97.3} \\
Predictor sharing & Shared / \textbf{Per-layer} & 96.4 / \textbf{97.3} \\
Calibration trajectories & 2 / 4 / \textbf{6} & 96.1 / 96.8 / \textbf{97.3} \\
\bottomrule
\end{tabular*}
\end{table}

\revised{Three denoising step groups yield 97.3\% success versus 95.5\% with one group. Ten groups yield 97.4\% with more predictor outputs, motivating the default $M=3$. For the uncertainty margin, $\alpha=3$ gives the highest observed success; increasing it to 5 broadens the scale range and yields 94.8\%.}

\revised{Full previous-chunk context yields 97.3\%, versus 95.9\% for the executed prefix and 96.3\% for the unexecuted suffix. Per-layer prediction yields 97.3\%, compared with 96.4\% for shared prediction. Calibration with two, four, and six trajectories yields 96.1\%, 96.8\%, and 97.3\%. The default uses two trajectories from each suite other than the evaluated suite.}

\subsection{Efficiency and Overhead Analysis}
\revised{We compare storage and memory traffic against QuantVLA's original MLP-only target, and measure AE latency on an NVIDIA A100 with batch size one.}

\begin{figure}[t]
\centering
\includegraphics[width=\linewidth]{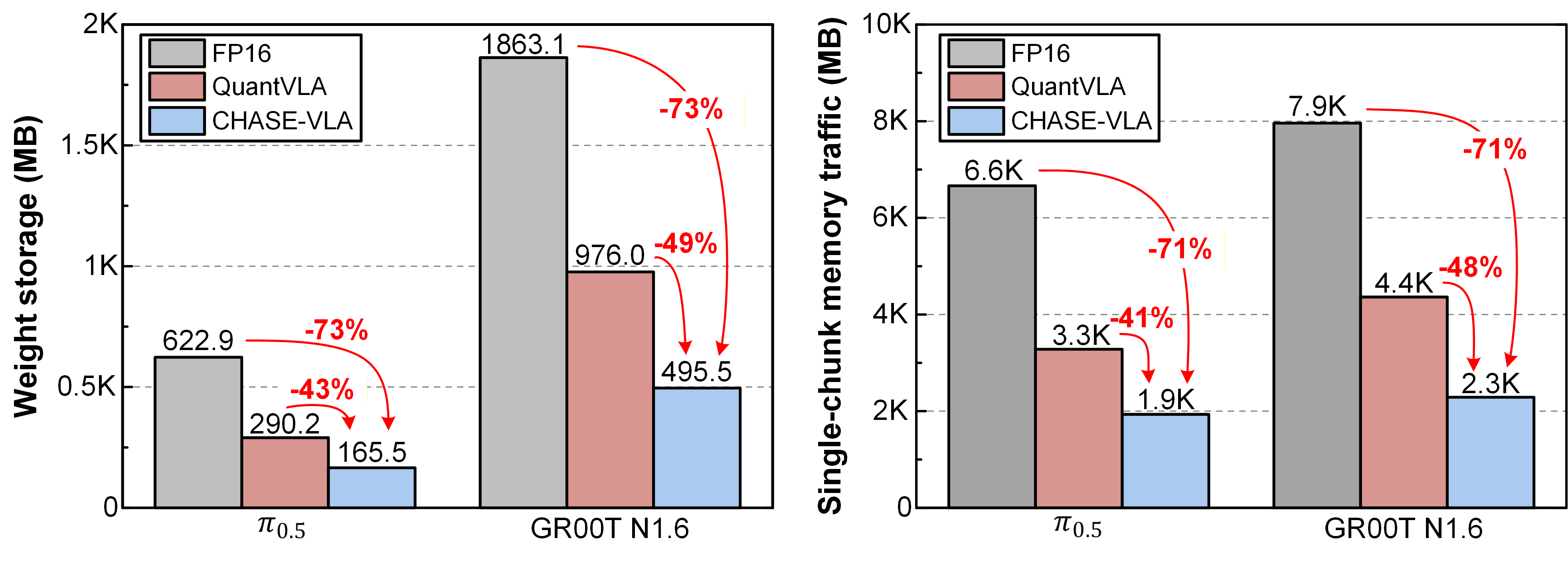}
\caption{Covered AE linear layer weight storage and single-chunk memory traffic on $\pi_{0.5}$ and GR00T N1.6, comparing FP16, QuantVLA, and CHASE-VLA.}
\label{fig:efficiency_graphs}
\end{figure}

\revised{Fig.~\ref{fig:efficiency_graphs} shows a 73.4\% reduction in quantized AE linear weight storage on both models, versus 53.4\% and 47.6\% for QuantVLA on $\pi_{0.5}$ and GR00T N1.6.}
\revised{Single-chunk memory traffic decreases by 70.9\% and 71.2\% relative to FP16, reaching 1935.55 MB and 2288.65 MB, respectively.}

\begin{table}[t]
\revisioncolor
\centering
\caption{\revised{End-to-end action expert latency and speedup on LIBERO, measured on an NVIDIA A100 GPU with batch size one. CHASE-VLA applies W4A4 quantization to all MLP and attention linear projections in the action expert. Speedups are computed relative to the corresponding FP16 baseline.}}
\label{tab:main_latency}
\small
\setlength{\tabcolsep}{3pt}
\begin{tabular*}{\linewidth}{@{\extracolsep{\fill}}lccccc@{}}
\toprule
\multirow{2}{*}{Model} & FP16 & \multicolumn{2}{c}{W4A8} & \multicolumn{2}{c}{CHASE-VLA (W4A4)} \\
\cmidrule(lr){2-2}\cmidrule(lr){3-4}\cmidrule(lr){5-6}
 & Latency (ms) & Latency (ms) & Speedup & Latency (ms) & Speedup \\
\midrule
$\pi_{0.5}$ & 26.957 & 25.585 & $1.054\times$ & 21.069 & $1.279\times$ \\
GR00T N1.6 & 43.003 & 41.889 & $1.027\times$ & 37.563 & $1.145\times$ \\
\bottomrule
\end{tabular*}
\end{table}

\revised{Table~\ref{tab:main_latency} reports end-to-end AE latency. CHASE-VLA reduces AE latency by 21.8\% on $\pi_{0.5}$ and 12.7\% on GR00T N1.6 relative to FP16, and by 17.7\% and 10.3\% relative to W4A8, respectively.}
\revised{These latency gains are accompanied by a small predictor storage overhead: 5.76 MB on $\pi_{0.5}$ and 11.50 MB on GR00T N1.6, corresponding to 1.26\% and 0.84\% of the saved weight storage, respectively.}

\section{Conclusion}
We introduced CHASE-VLA, a PTQ method that utilizes the generated action chunk as a causal signal for activation-scale estimation in diffusion-based VLA AEs.
By applying action-conditioned scaling at the denoising step group level, CHASE-VLA enables W4A4 quantization of both MLP and attention projections in the repeated AE without requiring modifications to the pretrained policy.
With W4A4 precision for AE linear layers, CHASE-VLA restores the FP16-level average LIBERO success on $\pi_{0.5}$ with 97.3\%, and maintains 90.5\% and 91.5\% average success on CogACT and GR00T N1.6.
It reduces the weight storage of the quantized AE linear layers by 73.4\% on both $\pi_{0.5}$ and GR00T N1.6, and decreases single-chunk memory traffic by 70.9\% and 71.2\%, with predictor overheads of only 1.26\% and 0.84\% of the saved storage, respectively.

%
%

\medskip
\noindent\textbf{Acknowledgments.}
This work was supported by the National Research Foundation of Korea (NRF) grants funded by the Korea government (MSIT) (RS-2025-02264052 and RS-2025-09322969), and by the Institute of Information \& Communications Technology Planning \& Evaluation (IITP) grant funded by the MSIT (RS-2026-25531289, Development of Incrementally Adaptive On-device Learning Technology for New Environments).

\clearpage
\bibliographystyle{splncs04}
\bibliography{main}


\renewcommand{\theHsection}{supp.\arabic{section}}
\renewcommand{\theHsubsection}{\theHsection.\arabic{subsection}}
\renewcommand{\theHsubsubsection}{\theHsubsection.\arabic{subsubsection}}
\renewcommand{\theHequation}{supp.\arabic{equation}}
\renewcommand{\theHfigure}{supp.\arabic{figure}}
\renewcommand{\theHtable}{supp.\arabic{table}}

\title{CHASE-VLA: Post-Training Quantization Framework for Vision-Language-Action Models with Chunk-Aware Scale Estimation\texorpdfstring{\\}{ }(Supplementary Material)}
\titlerunning{CHASE-VLA: PTQ Framework for VLA Models}

\author{Jin Hyun\inst{1,2}\orcidlink{0000-0002-4038-0170} \and
Jung Gyu Min\inst{1,2}\orcidlink{0009-0002-2041-617X} \and
Gyuhyun Jung\inst{1,2}\orcidlink{0009-0001-4826-537X} \and
Youngjoo Lee\inst{1}\orcidlink{0000-0002-2467-8276}}

\authorrunning{J.~Hyun et al.}

\institute{KAIST, Daejeon, South Korea\\
\email{youngjoo@kaist.ac.kr}
\and
POSTECH, Pohang, South Korea\\
\email{\{hyunj, mjg1104, jgh1020\}@postech.ac.kr}}

\maketitle

\section{Action Expert Architectures of VLA Models}
\label{sec:supp_model_structures}

Table~\ref{tab:supp_model_structures} summarizes the action expert (AE) architectures and repeated AE linear layers of Vision-Language-Action (VLA) models.

\begin{table}
\centering
\caption{AE architectures and quantized AE linear layers.}
\label{tab:supp_model_structures}
\small
\setlength{\tabcolsep}{3pt}
\begin{tabular}{@{\extracolsep{\fill}}lcccccc@{}}
\toprule
\multicolumn{1}{c}{\multirow[c]{2}{*}[-0.45ex]{Model}} &
\multirow[c]{2}{*}[-0.45ex]{\shortstack[c]{AE\\blocks}} &
\multirow[c]{2}{*}[-0.45ex]{\shortstack[c]{Denoising\\steps}} &
\multirow[c]{2}{*}[-0.45ex]{\shortstack[c]{Chunk\\$(H{\times}d_a)$}} &
\multirow[c]{2}{*}[-0.45ex]{\shortstack[c]{Target\\layers}} &
\multicolumn{2}{c}{Breakdown} \\
\cmidrule(lr){6-7}
\multicolumn{1}{c}{} & & & & & MLP & Attn. proj. \\
\midrule
$\pi_{0.5}$ & 18 & 10 & $50{\times}7$ & 126 & 54 & 72 \\
CogACT & 12 & 10 & $16{\times}7$ & 48 & 24 & 24  \\
GR00T N1.6 & 32 & 4 & $16{\times}7$ & 192 & 64 & 128  \\
\bottomrule
\end{tabular}
\end{table}

The chunk column reports the action chunk shape $H{\times}d_a$ used for CHASE-VLA chunk conditioning, where $H$ is the action horizon and $d_a$ is the action dimension.
The breakdown columns report the number of linear layers in each model's implementation: CogACT uses fused attention projections, whereas $\pi_{0.5}$ and GR00T N1.6 utilize separate attention projections.
For GR00T N1.6, the larger number of AE blocks and separate attention projections lead to a larger quantization target relative to $\pi_{0.5}$ and CogACT.

\section{CHASE-VLA Predictor Architecture}
\label{sec:supp_predictor_architecture}

CHASE-VLA uses a separate scale predictor for each quantized AE linear layer, since activation-scale statistics differ across the target linear layers.
Each predictor is stored in 16-bit floating point (FP16) and maps the action context through a two-layer MLP with a ReLU activation between the two linear layers, producing
\begin{equation}
  f_{\phi_l}(\operatorname{norm}(z))
  \in
  \mathbb{R}^{2M},
  \label{eq:supp_predictor_arch}
\end{equation}
where $z$ denotes the flattened previous action chunk of the corresponding VLA model, and $\operatorname{norm}(\cdot)$ denotes feature standardization.
The outputs are organized as one log-scale adjustment and one log-variance estimate for each denoising step group.
Table~\ref{tab:supp_predictor_architecture} summarizes the predictor count and the two-layer MLP dimensions used for each VLA model.

\begin{table}[t]
\centering
\caption{CHASE-VLA predictor architecture with a two-layer MLP.}
\label{tab:supp_predictor_architecture}
\small
\setlength{\tabcolsep}{3pt}
\begin{tabular}{@{\extracolsep{\fill}}lrrrrr@{}}
\toprule
Model & Predictors & $d_{\mathrm{in}}$ & $d_{\mathrm{hid}}$ & $M$ & $d_{\mathrm{out}}$ \\
\midrule
$\pi_{0.5}$ & 126 & 350 & 64 & 3 & 6 \\
CogACT & 48 & 112 & 256 & 3 & 6  \\
GR00T N1.6 & 192 & 112 & 256 & 2 & 4  \\
\bottomrule
\end{tabular}
\end{table}

In Table~\ref{tab:supp_predictor_architecture}, $d_{\mathrm{in}}$ denotes the flattened action-context dimension, and $d_{\mathrm{hid}}$ denotes the hidden width.
$M$ denotes the number of denoising step groups, and $d_{\mathrm{out}}=2M$ is the predictor output dimension.
The model-dependent value of $d_{\mathrm{in}}$ follows the action chunk shape in Table~\ref{tab:supp_model_structures}: $\pi_{0.5}$ uses $H=50$ with $d_a=7$, while CogACT and GR00T N1.6 use $H=16$ with $d_a=7$.

\section{CHASE-VLA Predictor Loss}
\label{sec:supp_loss}

CHASE-VLA fits its scale predictors as part of post-training calibration, with the pretrained VLA policy and AE weights kept frozen.
Only the predictor parameters $\phi_l$ are optimized on the calibration queries.
For each target linear layer $l$ and denoising step group $m$, the calibration target is the base-10 log-ratio between the current-chunk scale and the group-specific initial scale:
\begin{equation}
  y^{*}_{t_i,l,m}
  =
  \log_{10}
  \left(
  \epsilon+
  \frac{s^{\mathrm{chunk}}_{t_i,l,m}}{s^{\mathrm{init}}_{l,m}}
  \right).
  \label{eq:supp_scale_target}
\end{equation}
In Eq.~\ref{eq:supp_scale_target}, $t_i$ denotes the query time of the $i$-th collected calibration query, $m\in\{1,\ldots,M\}$ indexes denoising step groups, and $\epsilon$ is a small numerical constant.
In this target, $s^{\mathrm{init}}_{l,m}$ denotes the calibration-derived group-specific base scale used when no previous action chunk is available, while $s^{\mathrm{chunk}}_{t_i,l,m}$ denotes the scale measured from activations recorded during generation of the corresponding current chunk.

For each target linear layer $l$, the predictor takes the action context as input and outputs one log-scale adjustment and one log-variance estimate per denoising step group:
\begin{equation}
  \left[
  \hat{y}_{t_i,l,1:M},
  \log\hat{\sigma}^{2}_{t_i,l,1:M}
  \right]
  =
  f_{\phi_l}(\operatorname{norm}(z_{t_i})),
  \qquad
  z_{t_i}=\operatorname{vec}(A_{t_i-\tau}).
\end{equation}
The offset $\tau$ is the number of actions executed before the next policy query.
The residual in log-scale space is
\begin{equation}
  \Delta_{t_i,l,m}
  =
  y^{*}_{t_i,l,m}
  -
  \hat{y}_{t_i,l,m}.
\end{equation}
The predictor objective combines a pinball term, a heteroscedastic Gaussian negative log-likelihood (NLL) term for the log-variance head, and a mean-squared parameter penalty.
The pinball term is
\begin{equation}
  \mathcal{L}_{\mathrm{pin}}
  =
  \frac{1}{|\mathcal{D}_l|M}
  \sum_{i\in\mathcal{D}_l}
  \sum_{m=1}^{M}
  \max\left(
  \beta \Delta_{t_i,l,m},
  (\beta-1)\Delta_{t_i,l,m}
  \right).
  \label{eq:supp_pinball}
\end{equation}
In Eq.~\ref{eq:supp_pinball}, $\mathcal{D}_l$ is the set of collected calibration queries for which layer $l$ is observed, and $\beta$ is the quantile level.
Minimizing $\mathcal{L}_{\mathrm{pin}}$ fits a high quantile of the log-ratio target rather than its conditional mean, so that the predicted adjustment reflects large activation ranges that are critical under low-bit activation quantization.
The heteroscedastic NLL term is
\begin{equation}
  \mathcal{L}_{\mathrm{nll}}
  =
  \frac{1}{|\mathcal{D}_l|M}
  \sum_{i\in\mathcal{D}_l}
  \sum_{m=1}^{M}
  \left[
  \frac{1}{2}
  \exp(-\log\hat{\sigma}^{2}_{t_i,l,m})
  \Delta_{t_i,l,m}^{2}
  +
  \frac{1}{2}
  \log\hat{\sigma}^{2}_{t_i,l,m}
  \right].
  \label{eq:supp_nll}
\end{equation}
This term supervises the log-variance head in log-scale space; the predicted standard deviation, scaled by $\alpha$, provides the uncertainty margin used in the scale multiplier $\gamma$ in Eq.~\ref{eq:supp_inference_multiplier}.
The mean-squared parameter penalty is
\begin{equation}
  \mathcal{L}_{\mathrm{reg}}
  =
  \sum_{p\in\phi_l}
  \operatorname{mean}(p^2).
  \label{eq:supp_reg}
\end{equation}
This penalty regularizes the lightweight predictor and mitigates overfitting to the calibration queries.

The final predictor objective is
\begin{equation}
  \mathcal{L}_{\mathrm{CHASE}}
  =
  \mathcal{L}_{\mathrm{pin}}
  +
  \lambda
  \mathcal{L}_{\mathrm{nll}}
  +
  \eta
  \mathcal{L}_{\mathrm{reg}}.
  \label{eq:supp_total_loss}
\end{equation}
In the evaluation, we set $\beta=0.95$, $\lambda=0.5$, and $\eta=10^{-2}$.

At inference time, the predicted uncertainty is used only to form the scale multiplier applied to the group-specific initial scale:
\begin{equation}
  \hat{\sigma}_{t,l,m}
  =
  \exp\left(\frac{1}{2}\log\hat{\sigma}^{2}_{t,l,m}\right),
  \qquad
  \gamma_{t,l,m}
  =
  10^{\hat{y}_{t,l,m}+\alpha\hat{\sigma}_{t,l,m}},
  \label{eq:supp_inference_multiplier}
\end{equation}
where we set $\alpha=3.0$.

\section{CHASE-VLA Post-Training Quantization Details}
\label{sec:supp_calib}

\subsection{Calibration Setup}
In CHASE-VLA, post-training calibration estimates the quantization scales of the target AE linear layers and the activation-scale statistics used to fit the predictors.
Calibration is performed with the pretrained VLA policy and AE weights kept frozen throughout the procedure.
For each LIBERO evaluation suite, we construct a calibration set by randomly sampling two trajectories from each of the other LIBERO suites.
The evaluated suite is excluded from this sampling process, which prevents trajectory overlap between the calibration and evaluation sets.
The set is used only to estimate activation-scale statistics and to fit the scale predictors.

\subsection{Weight Quantization and Denoising Step Groups}
The quantization configuration is fixed from calibration through inference.
For the quantized AE linear layer weights, we apply symmetric per-row 4-bit quantization with the round-to-nearest method.
The denoising steps are partitioned once for each AE, and the resulting $M$ groups are reused throughout calibration and evaluation.
For $\pi_{0.5}$ and CogACT, which use $K=10$ denoising steps, the groups have sizes $3,4,$ and $3$ and correspond to step indices $\{0,1,2\}$, $\{3,4,5,6\}$, and $\{7,8,9\}$.
For GR00T N1.6, which uses $K=4$ denoising steps, the groups have sizes $2$ and $2$ and correspond to step indices $\{0,1\}$ and $\{2,3\}$.

\section{Confidence Intervals for LIBERO Success Rates}
\label{sec:supp_confidence_intervals}

Table~\ref{tab:supp_confidence_intervals} reports 95\% confidence intervals for the average LIBERO success rates presented in the main paper.
Expanded QuantVLA applies W4A4 quantization to both MLP and attention projections in the AE, matching the quantization target of CHASE-VLA.

\begin{table}[htbp]
\centering
\caption{95\% confidence intervals for average LIBERO success rates (\%).}
\label{tab:supp_confidence_intervals}
\small
\setlength{\tabcolsep}{4pt}
\begin{tabular*}{\linewidth}{@{\extracolsep{\fill}}lccc@{}}
\toprule
Method & $\pi_{0.5}$ & CogACT & GR00T N1.6 \\
\midrule
FP16 & [95.7, 98.1] & [91.2, 94.7] & [96.8, 98.8] \\
QuantVLA (expanded) & [93.6, 96.5] & [72.3, 78.2] & [45.9, 52.8] \\
Q-DiT & [92.6, 95.8] & [74.9, 80.6] & [20.1, 25.9] \\
PTQ4DiT & [91.6, 95.0] & [76.7, 82.3] & [17.4, 22.9] \\
CHASE-VLA & [95.9, 98.2] & [88.3, 92.3] & [89.4, 93.2] \\
\bottomrule
\end{tabular*}
\end{table}

On $\pi_{0.5}$, the CHASE-VLA interval overlaps the FP16 interval, and the point estimates remain close.
On CogACT and GR00T N1.6, the CHASE-VLA intervals are above and do not overlap those of the compared W4A4 PTQ baselines.

\section{\revised{Predictor-Free Scaling Baselines}}
\label{sec:supp_predictor_free}

\revised{Table~\ref{tab:supp_predictor_free} compares CHASE-VLA with three predictor-free scale-selection rules on $\pi_{0.5}$ LIBERO.}
\revised{Prev-query AbsMax reuses the previous query's activation maxima for each layer and denoising step group.}
\revised{AbsMax EMA smooths these maxima with an exponential moving average using $\rho=0.9$.}
\revised{Motion lookup maps the previous action chunk's energy and variation to calibrated scale tables.}

\begin{table}[htbp]
\revisioncolor
\centering
\caption{\revised{LIBERO success rates (\%) for predictor-free baselines and CHASE-VLA on $\pi_{0.5}$.}}
\label{tab:supp_predictor_free}
\small
\setlength{\tabcolsep}{3pt}
\begin{tabular*}{\linewidth}{@{\extracolsep{\fill}}lccccc@{}}
\toprule
Scale adjustment & Spatial & Object & Goal & Long & Avg. \\
\midrule
Prev-query AbsMax & 97.0 & 98.0 & 96.5 & 92.0 & 95.9 \\
AbsMax EMA & 97.5 & 98.5 & 95.5 & 92.5 & 96.0 \\
Motion lookup & 97.0 & 98.5 & 97.5 & 92.5 & 96.4 \\
CHASE-VLA & 97.0 & 99.5 & 98.0 & 94.5 & 97.3 \\
\bottomrule
\end{tabular*}
\end{table}

\revised{CHASE-VLA exceeds the strongest predictor-free baseline, motion lookup, by 0.9 percentage points in average success.}

\section{\revised{SimplerEnv Evaluation}}
\label{sec:supp_simpler}

\revised{Table~\ref{tab:supp_simpler} evaluates GR00T N1.6 on SimplerEnv under Variant Aggregation (VA) and Visual Matching (VM).}
\revised{For the quantized methods, the VLM remains in FP16 while the AE QKVO projections and MLP layers use W4A4.}
\revised{CHASE-VLA differs from FP16 by $+0.45$, $-2.67$, and $-1.72$ percentage points on Google Robot (VA), Google Robot (VM), and WidowX (VM), respectively.}

\begin{table}[htbp]
\revisioncolor
\centering
\caption{\revised{GR00T N1.6 success rates (\%) on SimplerEnv. VA denotes Variant Aggregation and VM denotes Visual Matching.}}
\label{tab:supp_simpler}
\small
\setlength{\tabcolsep}{4pt}
\begin{tabular*}{\linewidth}{@{\extracolsep{\fill}}lccc@{}}
\toprule
Evaluation setup & FP16 & QuantVLA & CHASE-VLA \\
\midrule
Google Robot (VA) & 46.54 & 29.47 & 46.99 \\
Google Robot (VM) & 56.00 & 43.67 & 53.33 \\
WidowX (VM) & 64.86 & 59.71 & 63.14 \\
\bottomrule
\end{tabular*}
\end{table}

\end{document}